\pdfoutput=1

\documentclass[11pt]{article}

\usepackage[backref=page]{hyperref}       

\usepackage{ACL2023}

\usepackage{times}
\usepackage{latexsym}

\usepackage[T1]{fontenc}

\usepackage[utf8]{inputenc}

\usepackage{microtype}

\usepackage{inconsolata}

\usepackage{xurl}
\usepackage{multirow}
\usepackage{subcaption}
\usepackage{array}

\usepackage{listings}
\usepackage{enumitem}
\usepackage{soul}
\usepackage{adjustbox}
\usepackage{float}
\usepackage{appendix}
\usepackage{booktabs}
\usepackage{tabularx}
\usepackage{arydshln}

\usepackage{xspace}

\makeatletter
\def\mathcolor#1#{\@mathcolor{#1}}
\def\@mathcolor#1#2#3{%
  \protect\leavevmode
  \begingroup
    \color#1{#2}#3%
  \endgroup
}
\definecolor{green}{rgb}{0.05, 0.9, 0.05}

\definecolor{redbw}{HTML}{d7191c}
\definecolor{greenbw}{HTML}{1c8036}
\definecolor{bluebw}{HTML}{2b83ba}

\newcommand\mup[1]{\footnotesize{\color{greenbw}#1\color{black}}}
\newcommand\mdo[1]{\footnotesize{\color{redbw}#1\color{black}}}
\newcommand\msa[1]{\footnotesize{#1}}

\newcommand{\name}{{\fontfamily{lmss}\selectfont Multi$\mathsf{^2}$SPE}\xspace}

\newcommand{\dataname}{{\fontfamily{lmss}\selectfont Multi-SciDocs}\xspace}

\newcommand{\VarCell}[2]{ \multicolumn{1}{#1}{\scriptsize $\pm$ #2} }

\usepackage{amsmath,amsfonts,bm}

\def\eqref#1{equation~\ref{#1}}

\def\1{\bm{1}}

\def\rmW{{\mathbf{W}}}

\def\vb{{\bm{b}}}
\def\vc{{\bm{c}}}

\def\vh{{\bm{h}}}

\DeclareMathAlphabet{\mathsfit}{\encodingdefault}{\sfdefault}{m}{sl}
\SetMathAlphabet{\mathsfit}{bold}{\encodingdefault}{\sfdefault}{bx}{n}

\def\gP{{\mathcal{P}}}

\usepackage{cleveref}

\title{Encoding Multi-Domain Scientific Papers \\ by Ensembling Multiple CLS Tokens}

\author{Ronald Seoh\thanks{* Equal contribution}\hspace{0.15cm}\thanks{\hspace{0.6em}Work done while at the University of Massachusetts Amherst.}\hspace{0.15cm}\textsuperscript{1} Haw-Shiuan Chang\footnotemark[1]\hspace{0.15cm}\footnotemark[2]\hspace{0.15cm}\textsuperscript{2} Andrew McCallum\textsuperscript{3}\\
\textsuperscript{1 }Purdue University \textsuperscript{2 }Amazon \textsuperscript{3 }University of Massachusetts Amherst \\
\texttt{bseoh@purdue.edu}, \texttt{chawshiu@amazon.com} \\
\texttt{mccallum@cs.umass.edu}\\}

\begin{document}
\maketitle
\begin{abstract}
Many useful tasks on scientific documents, such as topic classification and citation prediction, involve corpora that span multiple scientific domains. Typically, such tasks are accomplished by representing the text with a vector embedding obtained from a Transformer’s single CLS token. In this paper, we argue that using multiple CLS tokens could make a Transformer better specialize to multiple scientific domains. We present \name: it encourages each of multiple CLS tokens to learn diverse ways of aggregating token embeddings, then sums them up together to create a single vector representation. We also propose our new multi-domain benchmark, \dataname, to test scientific paper vector encoders under multi-domain settings. We show that \name reduces error by up to 25\% in multi-domain citation prediction, while requiring only a negligible amount of computation in addition to one BERT forward pass.
\end{abstract}

\section{Introduction}

With an ever-increasing amount of research publications, it has become virtually essential to develop NLP methods that would allow researchers to efficiently process the wealth of scientific knowledge. Leveraging pretrained language models and citation graphs, SPECTER~\citep{cohan2020specter} brings sizeable improvement over the previously state-of-the-art paper encoders and similarity estimation models such as SciBERT~\citep{beltagy2019scibert} and Citeomatic \citep{bhagavatula2018content}. Recently, SciNCL~\citep{ostendorff2022neighborhood} has introduced more sophisticated positive and negative sampling strategies to improve SPECTER further.

\begin{figure}[t]
\centering
\includegraphics[width=\columnwidth, clip, trim=23cm 6cm 16cm 5cm]{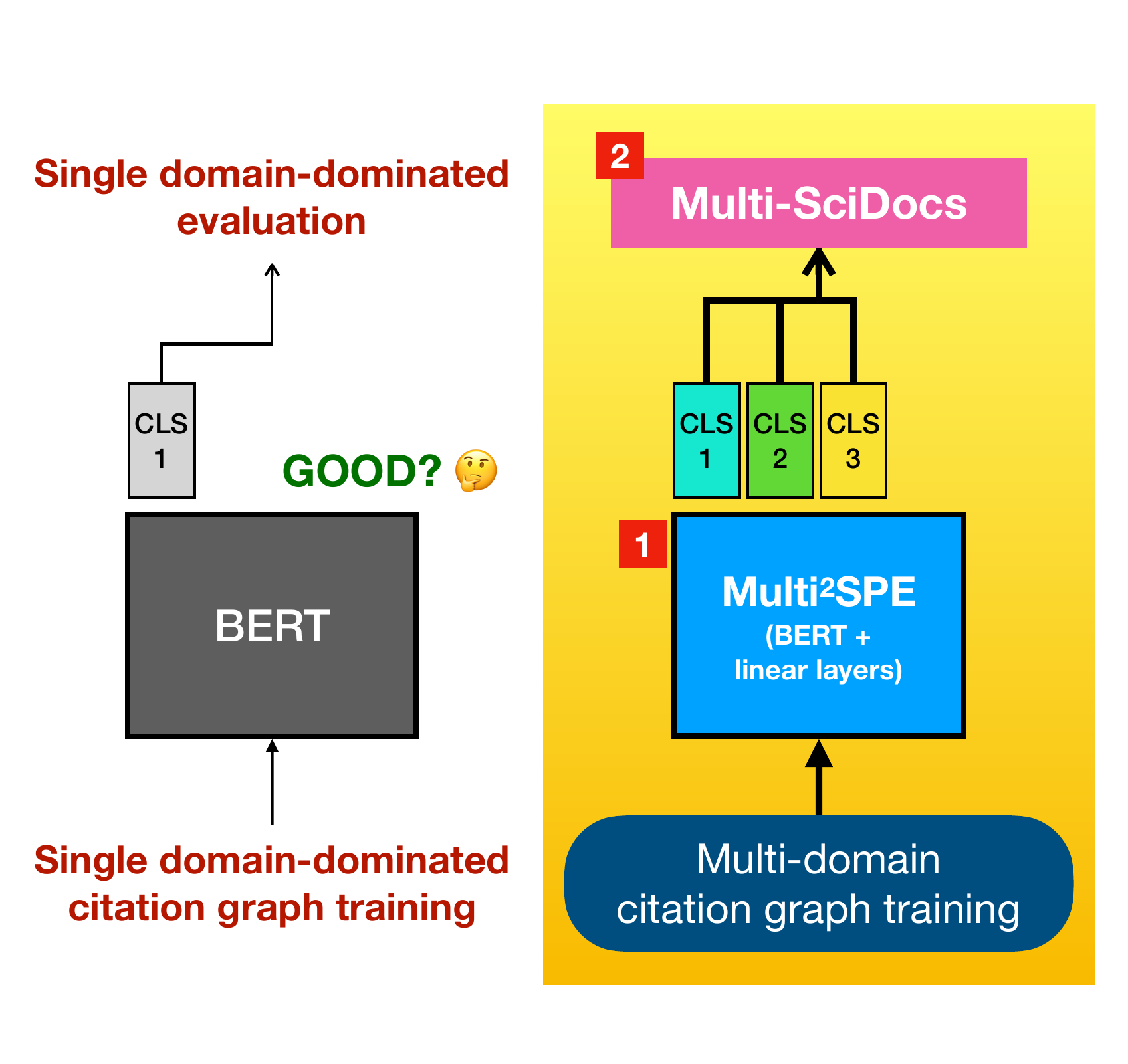}
\vspace*{-2.5em}
\caption{An overview of our two-parted solution. 1) \name 
better utilizes multi-domain citation data through multiple diversified CLS embeddings. 2) \dataname is our new benchmark for testing scientific papers embeddings under multi-domain settings.}
\label{fig:overview}
\end{figure}

Despite all the progress made so far, what is yet missing from literature is examining whether existing encoders can effectively represent the scientific papers across diverse subject areas.
In previous work \citep{bhagavatula2018content,beltagy2019scibert,cohan2020specter}, their training and evaluation data primarily consist of scientific papers from specific subject areas such as computer science and medicine. While these choices might be due to non-technical reasons such as the lack of open access articles \citep{piwowar2018state} or insufficient number of users from certain domains, it naturally makes us wonder whether 
we can represent papers from more diverse scientific domains using a single encoder, and whether we could improve state-of-the-art models under multi-domain settings. 

In this paper, we lay out our two-parted solution to overcome this limitation: the first part is our scientific paper encoder, \name. This is inspired by Multi-CLS BERT~\citep{MultiCLSBert} and built upon the intuition that extracting embeddings through just one CLS token is limiting, as more ideal ways of mixing contextualized word embeddings could be different for each subject areas. Instead, we add multiple CLS tokens to obtain embeddings that pay attention to different words, and ensemble the embeddings together to form a single paper representation. For the second part, we introduce the \dataname benchmark, to better understand the capabilities of scientific document representations in handling multi-domain settings.

Comparing \name and single CLS baselines on \dataname suggests that training \name on single domain-dominated citation graphs already boosts the scores on multi-domain tasks; with more balanced multi-domain training, \name brings even bigger improvements.

\section{\name: Multi-Domain \(\mathbf{\times}\) Multi-CLS Scientific Paper Encoder}
\label{sec:method}

We describe the major components of \name, our scientific paper encoder. 

Our key idea is to use multiple CLS tokens instead of just one: since one CLS embedding corresponds to merely a single scheme of aggregating word embeddings, it might be sufficient for the documents from one domain but may be far from ideal for the documents from other domains. We address this observation by prepending multiple CLS tokens to input documents and introducing small architectural additions that encourage each CLS embeddings to learn distinctive ways of mixing word embeddings together for the final document representation.

\subsection{Multiple CLS Encoder}
\label{sec:multi_CLS}

\begin{figure}[t!]
\centering
\includegraphics[width=1\linewidth]{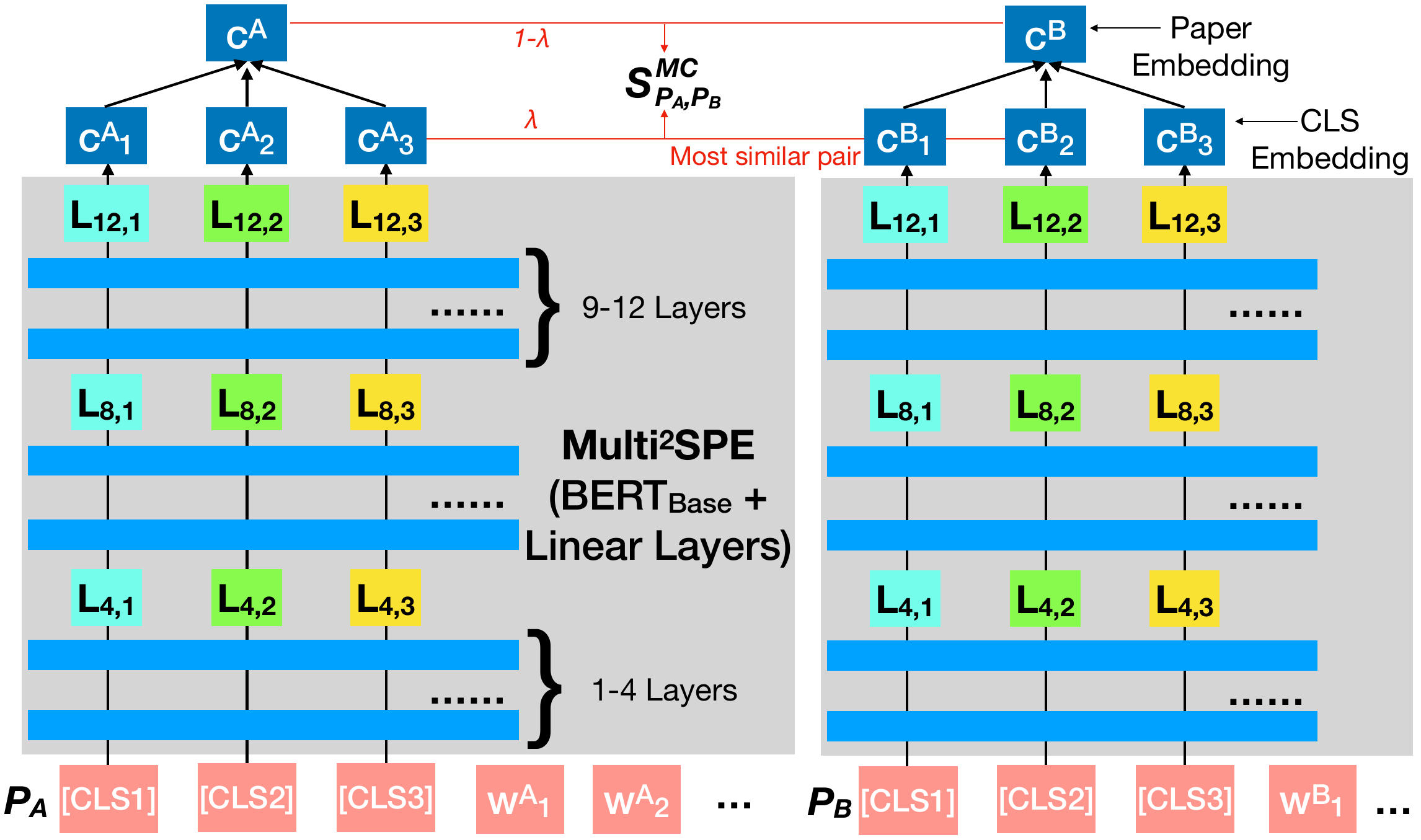}
\caption{The architecture of \name and its similarity measurement during training $\text{S}^{MC}_{\gP^A,\gP^B}$. }
\label{fig:contrastive_architecture}
\end{figure}

With multiple CLS tokens (\texttt{[CLS1], …, [CLSK]}), we insert linear layers $L_{l,k}$ at the sequence positions of each CLS embeddings as shown in \Cref{fig:contrastive_architecture}, to encourage the CLS embeddings to pay attention to different contextualized word embeddings. 

We use a re-parameterization trick to ensure that all the added linear transformations at each BERT layer are different and not similar to each other:

\vspace{-3mm}
\small
\begin{equation}
\label{eq:L_out_ft}
L_{l,k}(\vh^c_{l,k}) = (\rmW_{l,k} - \frac{1}{K} \sum_{k'} \rmW_{l,k'})\vh^c_{l,k} + \vb_{l,k},
\end{equation}
\normalsize

\noindent $L_{l,k}$ is the linear transformation for $k$th CLS token at the layer $l$, $\rmW_{l,k} - \frac{1}{K} \sum_{k'} \rmW_{l,k'}$ is the linear projection weights and $\vb_{l,k}$ is the bias term. To prevent $\rmW_{l,k} - \frac{1}{K} \sum_{k'} \rmW_{l,k'}=\mathbf{0}$, the gradient descent tend to learn different $\rmW_{l,k}$ for different $k$.

\subsection{Contrastive Citation Prediction Loss}
\label{sec:loss}

Existing state-of-the-art scientific paper encoders such as SPECTER~\citep{cohan2020specter} and SciNCL~\citep{ostendorff2022neighborhood} use training signals coming from a contrastive citation prediction task: their objective function is to encourage the embedding of each query paper to be close to those of the paper cited by them, and be far away the papers they did not cite, $\gP^{-}$.

Similarly, we minimize the cross entropy loss of a given query paper $\gP^{Q}$, a cited paper $\gP^{+}$, and a paper not cited, $\gP^{-}$:

\vspace{-3mm}
\small
\begin{align}
\label{eq:loss}
L_{\gP^{Q}, \gP^{+}, \gP^{-}} = 
- \log \left( \frac{ \exp(\text{S}^{MC}_{\gP^{Q}, \gP^{+}}) }{  \sum\limits_{\gP \in \{\gP^{+}, \gP^{-}\}  } \exp(\text{S}^{MC}_{\gP^{Q},{\gP}})} \right), 
\end{align}
\normalsize
\noindent where $\text{S}^{MC}_{\gP^{Q}, \gP^{+}}$ is the similarity between the query paper $\gP^{Q}$ and the cited paper $\gP^{+}$ from the multiple CLS encoder. It is also the logit score for predicting the paper $\gP^{+}$ as the cited paper.

\subsection{Measuring Document Similarity with Multiple Embeddings}
\label{sec:multi_sim}

One typical use of document embeddings is to perform a nearest neighbor search for retrieving candidates similar to the query document. While it would be possible to use each of CLS embeddings separately, or concatenate them together to encode each document, we would significantly increase the computational costs of the retrieval process. Instead, during inference, we simply take the summation of CLS embeddings from paper $A$ to be its final paper representation $\vc^{A} = \sum_k \vc_k^{A}$ and $\vc_k^{A} = L_{12,k}(\vh^{c,A}_{12,k})$.

During the contrastive training (\Cref{sec:loss}), we compute the similarity between two papers $\text{S}^{MC}_{\gP^A,\gP^B}$ using dot products between their paper embeddings $(\vc^{A})^T (\vc^{B})$ and the most similar CLS embeddings $\max_{i,j}(\vc_i^{A})^T \vc_j^{B}$:

\vspace{-3mm}
\small
\begin{align}
\label{eq:logit_multi}
\text{S}^{MC}_{\gP^A,\gP^B} = \lambda \max_{i,j} (\vc_i^{A})^T \vc_j^{B} + 
(1-\lambda) (\vc^{A})^T (\vc^{B}),
\end{align}
\normalsize

\noindent where $\vc^{A} = \sum_k \vc_k^{A}$ and $\lambda$ is the hyperparameter for controlling the dependency between the CLS embeddings. Smaller $\lambda$ makes similarity measurement in training and testing more consistent and encourages the CLS embeddings to collaborate with each other. Larger $\lambda$ encourages each of the CLS embeddings to become more meaningful paper embeddings on their own.

\section{\dataname}
\label{sec:multi_scidocs}

\citet{cohan2020specter} proposed SciDocs as a comprehensive benchmark for evaluating scientific paper embeddings. SciDocs introduces 12 metrics from 7 tasks, but we have discovered that the domain distributions of 5 tasks are heavily biased toward computer science (CS) papers.\footnote{Please see \Cref{sec:dataset_stat} for detailed statistics.} The only exceptions are MeSH (Medical Subject Headings)~\cite{lipscomb2000medical}, which covers the papers from the biomedical domain, and MAG (Microsoft Academic Graph)~\citep{sinha2015overview}, which is a document classification task into 19 subject areas.

Thus, for a better measurement of multi-domain performance, we have created the multi-domain (co-)citation prediction tasks. We refer to the collection of 3 multi-domain tasks, \textsf{multi. cite}, \textsf{multi. co-cite}, and \textsf{MAG} as \dataname.

For \textsf{multi. (co-)cite} datasets, we randomly sample the query papers from S2ORC~\citep{lo2020s2orc}, avoiding a certain domain from being the majority of query papers, and follow the construction procedure of \textsf{(co-)cite} in SciDocs to get the positive and negative papers. 
For each query, we collect 500 negative papers and up to 5 positive papers. The task is to assign higher similarity scores to the positive papers and lower scores to the negative papers. In both datasets, the negative samples come from randomly sampled papers. In the \textsf{multi. cite} dataset, the positive samples are the papers cited by the query paper. In the \textsf{multi. co-cite} dataset, the positive samples, and the query paper are both cited by another paper.

\section{Experiments and Analyses}
\label{sec:experiments}

\begin{table*}[ht]
    \centering
    \begin{subtable}[t]{0.46\textwidth}
    \begin{adjustbox}{width=\textwidth}
    \begin{tabular}{c|c|cc|cc|c}
                                             & \textbf{MAG}  & \multicolumn{2}{c|}{\textbf{multi. cite}} & \multicolumn{2}{c|}{\textbf{multi. co-cite}} & \multirow{2}{*}{\textbf{Avg}}\\
                                             & \msa{F1}      & \msa{MAP}           & \msa{nDCG}         & \msa{MAP}          & \msa{nDCG}                  \\
        \midrule
        {\centering SPECTER}                   & 78.90                & 78.14         & 88.06         & 65.97        & 73.46        & 76.90 \\
        \hdashline
                                & 80.24                & 81.57         & 90.12         & 69.12        & 75.97        & 79.40 \vspace{-0.2cm} \\
        {\centering \name} & \VarCell{r|}{0.18} \vspace{-0.1cm}& \VarCell{r}{0.13} & \VarCell{r|}{0.07} & \VarCell{r}{0.08} & \VarCell{r|}{0.07} & \VarCell{r}{0.04}  \\
        {(3 CLS, \(\lambda=0.1\))}              & \mup{6.35\%}   & \mup{15.71\%} & \mup{17.29\%} & \mup{9.25\%} & \mup{9.45\%} & \mup{10.82\%} \\
        \specialrule{.2em}{.3em}{.3em}
        {\centering SciNCL}                       & 79.59                & 82.45         & 90.56         & 69.94        & 76.62        & 79.83 \\
        \hdashline
                                & 80.73                & 83.25         & 91.05         & 71.10        & 77.51        & 80.73 \vspace{-0.2cm} \\
        {\centering \name} & \VarCell{r|}{0.27} \vspace{-0.1cm}& \VarCell{r}{0.21} & \VarCell{r|}{0.12} & \VarCell{r}{0.32} & \VarCell{r|}{0.24} & \VarCell{r}{0.19}  \\
        {(3 CLS, \(\lambda=0.1\))}              & \mup{5.58\%} & \mup{4.57\%} & \mup{5.16\%} & \mup{3.86\%} & \mup{3.80\%} & \mup{4.44\%} \\
        
        \bottomrule
    \end{tabular}
    \end{adjustbox}
    \caption{Single domain (CS) training}
    \end{subtable}
    \hspace{\fill}
    \begin{subtable}[t]{0.46\textwidth}
    \begin{adjustbox}{width=\textwidth}
    \begin{tabular}{c|c|cc|cc|c}
                                             & \textbf{MAG}  & \multicolumn{2}{c|}{\textbf{multi. cite}} & \multicolumn{2}{c|}{\textbf{multi. co-cite}} & \multirow{2}{*}{\textbf{Avg}}\\
                                             & \msa{F1}                      & \msa{MAP}           & \msa{nDCG}          & \msa{MAP}          & \msa{nDCG}                  \\
        \midrule
        {\centering SPECTER}                      & 79.99              & 79.30         & 88.73         & 68.59        & 75.60        & 77.97 \\
        \hdashline
                                & 81.36               & 84.08         & 91.54         & 71.79        & 78.15        & 81.10 \vspace{-0.2cm} \\
        {\centering \name} & \VarCell{r|}{0.29} \vspace{-0.1cm}& \VarCell{r}{0.10} & \VarCell{r|}{0.06} & \VarCell{r}{0.22} & \VarCell{r|}{0.17} & \VarCell{r}{0.06}  \\
        {(3 CLS, \(\lambda=0.1\))}              & \mup{6.85\%}  & \mup{23.08\%} & \mup{24.96\%} & \mup{10.18\%} & \mup{10.45\%} & \mup{14.20\%} \\
        \specialrule{.2em}{.3em}{.3em}
        {\centering SciNCL}                       & 80.27               & 85.18         & 92.15         & 73.02        & 79.08        & 81.94 \\
        \hdashline

                                & 81.04               & 85.73         & 92.46         & 74.05        & 79.85        & 82.63 \vspace{-0.2cm} \\
        {\centering \name} & \VarCell{r|}{0.05} \vspace{-0.1cm}& \VarCell{r}{0.29} & \VarCell{r|}{0.17} & \VarCell{r}{0.25} & \VarCell{r|}{0.19} & \VarCell{r}{0.18}  \\
        {(3 CLS, \(\lambda=0.1\))}              & \mup{3.90\%}  & \mup{3.73\%} & \mup{3.98\%} & \mup{3.82\%} & \mup{3.69\%} & \mup{3.81\%} \\
        \bottomrule
    \end{tabular}
    \end{adjustbox}
    \caption{Multiple domain training}
    \end{subtable}
    \caption{Results of our methods and baselines on \dataname. All scores are averaged over four random seeds. We show standard errors as their confidence intervals. Percentages indicate relative error reduction over the baselines (SPECTER or SciNCL), which is an important metric for the models with high accuracy.}
    \label{tab:results}
\end{table*}

In the experiments, we evaluate \name and the corresponding baselines with \dataname. SPECTER and SciNCL are our single [CLS] token baselines: both use identical neural architectures and loss functions, but differ in sampling methods used to create their contrastive triples. Since we found training datasets in previous literature to be potentially limiting in handling papers from various scientific domains, we build our own multi-domain training datasets that follow the same sampling methods of SPECTER and SciNCL, but are more balanced in terms of the domain distribution.\footnote{Please see \Cref{sec:dataset_stat} for the comparison of domain distribution of SPECTER/SciNCL single-domain datasets and our multi-domain datasets.}

\subsection{Results}

Our main results are shown in \Cref{tab:results}. We can see that \name has consistently outperformed the baselines in all training cases. 
The scores from \textsf{MAG} show that \name is better capable of classifying the texts into diverse subject domains. In multi-domain citation prediction, its error reductions are up to 25\%. 
We hypothesize that the large improvement partially comes from the prevalent cross-domain citations in both our training and evaluation data. We note that the overall gains are smaller for SciNCL. We suspect that SciNCL's sampling method reduces the number of cross-domain citations in the dataset, which would have helped increase the diversity in CLS embeddings.

\subsection{Ablation Studies}

\begin{table}[h]
    \centering
    \begin{adjustbox}{width=0.95\columnwidth}
    \begin{tabular}{p{0.8\linewidth}ccc}
                                                             & \textbf{MAG} & \textbf{cite}      & \textbf{co-cite} \\
                                                             & \msa{F1}     & \msa{MAP}          & \msa{MAP} \\
        \midrule
        {\name (3 CLS, \(\lambda=0.1\))}                       & 81.36        & 84.08              & 71.79           \\
        \specialrule{.2em}{.3em}{.3em}
        \textbf{CLS Embedding Independence} & & \\
        \multirow{2}{*}{\(\rightarrow \lambda=0.0\)}         & 81.06         & 84.09              & 71.90 \\
                                                             & \mdo{-1.64\%} & \mup{0.11\%}       & \mup{0.39\%}\\
        \midrule
        \multirow{2}{*}{\(\rightarrow \lambda=0.5\)}         & 80.70         & 83.05              & 71.46 \\
                                                             & \mdo{-3.53\%} & \mdo{-6.45\%}      & \mdo{-1.18\%}\\
        \midrule
        \multirow{2}{*}{\(\rightarrow \lambda=1.0\)}         & 79.78         & 83.69              & 71.18 \\
                                                             & \mdo{-8.49\%} & \mdo{-2.40\%}      & \mdo{-2.15\%}\\
        \specialrule{.15em}{.3em}{.3em}
        \textbf{Architectural Changes} & & \\
        \multirow{2}{*}{\(\rightarrow\) 1 CLS token}         & 80.32         & 83.32              & 70.30 \\
                                                             & \mdo{-5.57\%} & \mdo{-4.76\%}      & \mdo{-5.30\%} \\
        \midrule
        \multirow{2}{*}{\(\rightarrow\) 5 CLS tokens}        & 80.83         & 84.03              & 72.59 \\
                                                             & \mdo{-2.84\%} & \mdo{-0.30\%}      & \mup{2.84\%} \\
        \midrule
        \multirow{2}{*}{\(\rightarrow\) No linear layer injection in BERT}   & 80.82         & 83.54            & 71.12 \\
                                                                             & \mdo{-2.88\%} & \mdo{-3.38\%}    & \mdo{-2.39\%}\\
        \midrule
        \multirow{2}{*}{\(\rightarrow\) No re-parameterization trick}        & 80.89         & 83.79            & 71.11 \\
                                                                             & \mdo{-2.55\%} & \mdo{-1.82\%}    & \mdo{-2.42\%}\\
        \bottomrule
    \end{tabular}
    \end{adjustbox}
    \caption{Ablation studies conducted on SPECTER and multiple domain training data. All scores are averaged over four random seeds. Percentages indicate relative error reduction over the baseline (3 CLS, \(\lambda=0.1\)).}
    \label{tab:ablation}
\end{table}

In \autoref{tab:ablation}, we start by examining the effect of \(\lambda\), the hyperparameter for controlling dependencies between CLS embeddings. While the differences are relatively small for \(\lambda=0.0\), we observe noticeable performance drops as we increase \(\lambda\) to 0.5 and 1.0. Our intuition is that it is generally more beneficial to encourage all embeddings to become a meaningful whole together, rather than directing each of them to stand on their own.

In the second set of our ablation studies, we quantify the performance benefits of each architectural changes we introduced in \Cref{sec:multi_CLS}. We can see that multiple CLS tokens are crucial, as having just one CLS leads to clear performance drop. 
Increasing the number of CLS tokens from 3 to 5 leads to mixed results. Their overall similar performance suggests that the quality of our multi-domain paper embeddings is not sensitive to the number of CLS tokens.
Lastly, we observe that both the linear layer injection to BERT and the re-parameterization trick have clear contributions to our models' better performance.

\section{Related Work}

Many studies focus only specific scientific NLP tasks such as citation recommendation~\citep{bhagavatula2018content,farber2020citation,farber2020hybridcite,ma2020review} and paper recommendation~\citep{beel2016paper, zhang2020multi}. Instead, our goal is improving upon general-purpose scientific paper encoders such as SPECTER~\citep{cohan2020specter} and SciNCL~\citep{ostendorff2022neighborhood}.

Another line of efforts relies on pre-defined facets~\citep{chakraborty2016ferosa,chan2018solvent,ostendorff2020aspect} or topics~\citep{zhang2020multi} for specific domains of interest, and measure paper similarities based on those facets/topics. Recently, \citet{mysore2021multi} suggests encoding a paper into multiple sentence embeddings to allow the users to search similar papers using partially constructed query papers. In contrast, \name automatically learns to identify the facets that are helpful for the citation prediction task, and combines all the facets into a single embedding to improve the similarity measurement and nearest neighbor search over the single CLS baseline while maintaining similar computational costs.

\citet{MultiCLSBert} proposes an efficient BERT ensemble model called Multi-CLS BERT, which inserts different linear layers for different CLS tokens, and re-parameterize its top linear layer during fine-tuning on the tasks in GLUE~\citep{wang2018glue} and SuperGLUE~\citep{wang2019superglue}. As explained in \autoref{sec:method}, we also envisioned that multiple CLS tokens could be  beneficial, especially for our case as all the scientific papers would come from a wide range of distinctive disciplines.
However, there are key differences between Multi-CLS BERT and our encoder, such as

using the re-parameterization trick in all the inserted linear layers during contrastive learning, and summing all the unnormalized CLS embeddings into a single paper embedding for the scientific paper similarity tasks. 

Moreover, we find that such efficient ensembling is especially beneficial when being trained and tested using multi-domain papers.

\section{Conclusion}

In this work, we identified insufficiencies in existing training datasets, evaluation benchmarks, and encoder architectures, when handling diverse subject domains in scientific literature. To overcome the current limitations, we introduced \name, a modified BERT encoder that learns a diversified set of embeddings from multi-domain citation data, and \dataname, our new benchmark for testing the embeddings of scientific papers using multi-domain tasks. Our experiments show that \name provides consistent improvements over the SOTA baselines. The ablation studies confirm the effectiveness of our modifications to BERT.

\section*{Limitations}

While we find that the training and evaluation datasets we have created allow us to better represent different scientific domains, how we could treat all subject domains more fairly in general is very much an open problem—especially with many real-life constraints on scientific literatures such as the scarcity of open access articles in certain domains. The domain distribution\footnote{\url{https://github.com/allenai/s2orc}} of S2ORC \citep{lo2020s2orc}, our primary source of citation records and full texts, suggests that there is apparent imbalance in the number of papers available across different domains. This situation gets even more complicated when we consider the fact that certain areas are strongly related to each other due to the nature of their subjects (e.g., Mathematics/Computer Science, Medicine/Biology.)

We believe that the three tasks chosen in \dataname are discriminative and objective measurements of the encoder's potential ability to handle multi-domain settings. However, it is yet to be proven that the high correlation exists between these intrinsic evaluation metrics and the actual effectiveness of any real-life systems that utilize paper representation methods.

Lastly, there is more progress to be made to achieve a complete understanding of how \name brings performance improvements. More specifically, we would like to perform more qualitative investigation on the specific roles of each CLS embeddings and how they collaborate with each other to create a single representation.

\section*{Ethical and Broader Impact}
\label{impact}

We believe that SOTA NLP techniques could deliver healthy boosts in the productivity and creativity of any researchers, regardless of their academic disciplines. For example, better similarity measurements between the papers across multiple domains could improve scientific paper retrieval systems: this would allow researchers to efficiently navigate through the wealth of scientific knowledge, and could eventually serve a significant role in encouraging new research activities.

\section*{Acknowledgements}
We thank Purujit Goyal for discovering a critical bug in our code.
This work was supported 
in part by the Center for Data Science and the Center for Intelligent Information Retrieval, 
in part by the Chan Zuckerberg Initiative under the project Scientific Knowledge Base Construction, 
in part by the IBM Research AI through the AI Horizons Network, 
in part using high-performance computing equipment obtained under a grant from the Collaborative R\&D Fund managed by the Massachusetts Technology Collaborative, 
and in part by the National Science Foundation (NSF) grant numbers IIS-1922090 and IIS-1763618.
Any opinions, findings, conclusions, or recommendations expressed in this material are those of the authors and do not necessarily reflect those of the sponsor.

\bibliography{anthology,custom}

\begin{thebibliography}{23}
\expandafter\ifx\csname natexlab\endcsname\relax\def\natexlab#1{#1}\fi

\bibitem[{Beel et~al.(2016)Beel, Gipp, Langer, and Breitinger}]{beel2016paper}
Joeran Beel, Bela Gipp, Stefan Langer, and Corinna Breitinger. 2016.
\newblock paper recommender systems: a literature survey.
\newblock \emph{International Journal on Digital Libraries}, 17(4):305--338.

\bibitem[{Beltagy et~al.(2019)Beltagy, Cohan, and Lo}]{beltagy2019scibert}
Iz~Beltagy, Arman Cohan, and Kyle Lo. 2019.
\newblock {SciBERT}: Pretrained contextualized embeddings for scientific text.
\newblock In \emph{EMNLP}.

\bibitem[{Bhagavatula et~al.(2018)Bhagavatula, Feldman, Power, and
  Ammar}]{bhagavatula2018content}
Chandra Bhagavatula, Sergey Feldman, Russell Power, and Waleed Ammar. 2018.
\newblock \href {https://doi.org/10.18653/v1/N18-1022} {Content-based citation
  recommendation}.
\newblock In \emph{Proceedings of the 2018 Conference of the North {A}merican
  Chapter of the Association for Computational Linguistics: Human Language
  Technologies, Volume 1 (Long Papers)}, pages 238--251, New Orleans,
  Louisiana. Association for Computational Linguistics.

\bibitem[{Chakraborty et~al.(2016)Chakraborty, Krishna, Singh, Ganguly, Goyal,
  and Mukherjee}]{chakraborty2016ferosa}
Tanmoy Chakraborty, Amrith Krishna, Mayank Singh, Niloy Ganguly, Pawan Goyal,
  and Animesh Mukherjee. 2016.
\newblock {FeRoSA}: A faceted recommendation system for scientific articles.
\newblock In \emph{Pacific-Asia Conference on Knowledge Discovery and Data
  Mining}.

\bibitem[{Chan et~al.(2018)Chan, Chang, Hope, Shahaf, and
  Kittur}]{chan2018solvent}
Joel Chan, Joseph~Chee Chang, Tom Hope, Dafna Shahaf, and Aniket Kittur. 2018.
\newblock {SOLVENT}: A mixed initiative system for finding analogies between
  research papers.
\newblock \emph{Proceedings of the ACM on Human-Computer Interaction},
  2(CSCW):1--21.

\bibitem[{Chang et~al.(2022)Chang, Sun, Ricci, and McCallum}]{MultiCLSBert}
Haw-Shiuan Chang, Ruei-Yao Sun, Kathryn Ricci, and Andrew McCallum. 2022.
\newblock \href {https://arxiv.org/abs/2210.05043} {{Multi-CLS BERT}: An
  efficient alternative to traditional ensembling}.
\newblock \emph{arXiv preprint arXiv:2210.05043}.

\bibitem[{Cohan et~al.(2020)Cohan, Feldman, Beltagy, Downey, and
  Weld}]{cohan2020specter}
Arman Cohan, Sergey Feldman, Iz~Beltagy, Doug Downey, and Daniel Weld. 2020.
\newblock \href {https://doi.org/10.18653/v1/2020.acl-main.207} {{SPECTER}:
  Document-level representation learning using citation-informed transformers}.
\newblock In \emph{Proceedings of the 58th Annual Meeting of the Association
  for Computational Linguistics}, pages 2270--2282, Online. Association for
  Computational Linguistics.

\bibitem[{Falcon et~al.(2020)Falcon, Borovec, Wälchli, Eggert, Schock, Jordan,
  Skafte, Ir1dXD, Bereznyuk, Harris, Murrell, Yu, Præsius, Addair, Zhong,
  Lipin, Uchida, Bapat, Schröter, Dayma, Karnachev, Kulkarni, Komatsu,
  Martin.B, SCHIRATTI, Mary, Byrne, Eyzaguirre, cinjon, and
  Bakhtin}]{william_falcon_2020_3828935}
William Falcon, Jirka Borovec, Adrian Wälchli, Nic Eggert, Justus Schock,
  Jeremy Jordan, Nicki Skafte, Ir1dXD, Vadim Bereznyuk, Ethan Harris, Tullie
  Murrell, Peter Yu, Sebastian Præsius, Travis Addair, Jacob Zhong, Dmitry
  Lipin, So~Uchida, Shreyas Bapat, Hendrik Schröter, Boris Dayma, Alexey
  Karnachev, Akshay Kulkarni, Shunta Komatsu, Martin.B, Jean-Baptiste
  SCHIRATTI, Hadrien Mary, Donal Byrne, Cristobal Eyzaguirre, cinjon, and Anton
  Bakhtin. 2020.
\newblock \href {https://doi.org/10.5281/zenodo.3828935}
  {Pytorchlightning/pytorch-lightning: 0.7.6 release}.

\bibitem[{F{\"a}rber and Jatowt(2020)}]{farber2020citation}
Michael F{\"a}rber and Adam Jatowt. 2020.
\newblock \href {https://arxiv.org/abs/2002.06961} {Citation recommendation:
  Approaches and datasets}.
\newblock \emph{ArXiv preprint}, abs/2002.06961.

\bibitem[{F{\"a}rber and Sampath(2020)}]{farber2020hybridcite}
Michael F{\"a}rber and Ashwath Sampath. 2020.
\newblock Hybridcite: A hybrid model for context-aware citation recommendation.
\newblock In \emph{Proceedings of the ACM/IEEE Joint Conference on Digital
  Libraries in 2020}, pages 117--126.

\bibitem[{Lipscomb(2000)}]{lipscomb2000medical}
Carolyn~E Lipscomb. 2000.
\newblock Medical subject headings {(MeSH)}.
\newblock \emph{Bulletin of the Medical Library Association}, 88(3):265.

\bibitem[{Lo et~al.(2020)Lo, Wang, Neumann, Kinney, and Weld}]{lo2020s2orc}
Kyle Lo, Lucy~Lu Wang, Mark Neumann, Rodney Kinney, and Daniel Weld. 2020.
\newblock \href {https://doi.org/10.18653/v1/2020.acl-main.447} {{S}2{ORC}: The
  semantic scholar open research corpus}.
\newblock In \emph{Proceedings of the 58th Annual Meeting of the Association
  for Computational Linguistics}, pages 4969--4983, Online. Association for
  Computational Linguistics.

\bibitem[{Ma et~al.(2020)Ma, Zhang, and Liu}]{ma2020review}
Shutian Ma, Chengzhi Zhang, and Xiaozhong Liu. 2020.
\newblock A review of citation recommendation: from textual content to enriched
  context.
\newblock \emph{Scientometrics}, pages 1--28.

\bibitem[{Mysore et~al.(2022)Mysore, Cohan, and Hope}]{mysore2021multi}
Sheshera Mysore, Arman Cohan, and Tom Hope. 2022.
\newblock Multi-vector models with textual guidance for fine-grained scientific
  document similarity.
\newblock In \emph{NAACL}.

\bibitem[{Ostendorff et~al.(2022)Ostendorff, Rethmeier, Augenstein, Gipp, and
  Rehm}]{ostendorff2022neighborhood}
Malte Ostendorff, Nils Rethmeier, Isabelle Augenstein, Bela Gipp, and Georg
  Rehm. 2022.
\newblock \href {https://arxiv.org/abs/2202.06671} {Neighborhood contrastive
  learning for scientific document representations with citation embeddings}.
\newblock \emph{ArXiv preprint}, abs/2202.06671.

\bibitem[{Ostendorff et~al.(2020)Ostendorff, Ruas, Blume, Gipp, and
  Rehm}]{ostendorff2020aspect}
Malte Ostendorff, Terry Ruas, Till Blume, Bela Gipp, and Georg Rehm. 2020.
\newblock \href {https://doi.org/10.18653/v1/2020.coling-main.545}
  {Aspect-based document similarity for research papers}.
\newblock In \emph{Proceedings of the 28th International Conference on
  Computational Linguistics}, pages 6194--6206, Barcelona, Spain (Online).
  International Committee on Computational Linguistics.

\bibitem[{Paszke et~al.(2019)Paszke, Gross, Massa, Lerer, Bradbury, Chanan,
  Killeen, Lin, Gimelshein, Antiga, Desmaison, K{\"{o}}pf, Yang, DeVito,
  Raison, Tejani, Chilamkurthy, Steiner, Fang, Bai, and
  Chintala}]{NEURIPS2019_9015}
Adam Paszke, Sam Gross, Francisco Massa, Adam Lerer, James Bradbury, Gregory
  Chanan, Trevor Killeen, Zeming Lin, Natalia Gimelshein, Luca Antiga, Alban
  Desmaison, Andreas K{\"{o}}pf, Edward Yang, Zachary DeVito, Martin Raison,
  Alykhan Tejani, Sasank Chilamkurthy, Benoit Steiner, Lu~Fang, Junjie Bai, and
  Soumith Chintala. 2019.
\newblock \href
  {https://proceedings.neurips.cc/paper/2019/hash/bdbca288fee7f92f2bfa9f7012727740-Abstract.html}
  {Pytorch: An imperative style, high-performance deep learning library}.
\newblock In \emph{Advances in Neural Information Processing Systems 32: Annual
  Conference on Neural Information Processing Systems 2019, NeurIPS 2019,
  December 8-14, 2019, Vancouver, BC, Canada}, pages 8024--8035.

\bibitem[{Piwowar et~al.(2018)Piwowar, Priem, Larivi{\`e}re, Alperin, Matthias,
  Norlander, Farley, West, and Haustein}]{piwowar2018state}
Heather Piwowar, Jason Priem, Vincent Larivi{\`e}re, Juan~Pablo Alperin, Lisa
  Matthias, Bree Norlander, Ashley Farley, Jevin West, and Stefanie Haustein.
  2018.
\newblock The state of oa: a large-scale analysis of the prevalence and impact
  of open access articles.
\newblock \emph{PeerJ}, 6:e4375.

\bibitem[{Sinha et~al.(2015)Sinha, Shen, Song, Ma, Eide, Hsu, and
  Wang}]{sinha2015overview}
Arnab Sinha, Zhihong Shen, Yang Song, Hao Ma, Darrin Eide, Bo-June Hsu, and
  Kuansan Wang. 2015.
\newblock An overview of microsoft academic service (mas) and applications.
\newblock In \emph{Proceedings of the 24th international conference on world
  wide web}, pages 243--246.

\bibitem[{Wang et~al.(2019{\natexlab{a}})Wang, Pruksachatkun, Nangia, Singh,
  Michael, Hill, Levy, and Bowman}]{wang2019superglue}
Alex Wang, Yada Pruksachatkun, Nikita Nangia, Amanpreet Singh, Julian Michael,
  Felix Hill, Omer Levy, and Samuel~R. Bowman. 2019{\natexlab{a}}.
\newblock \href
  {https://proceedings.neurips.cc/paper/2019/hash/4496bf24afe7fab6f046bf4923da8de6-Abstract.html}
  {Superglue: {A} stickier benchmark for general-purpose language understanding
  systems}.
\newblock In \emph{Advances in Neural Information Processing Systems 32: Annual
  Conference on Neural Information Processing Systems 2019, NeurIPS 2019,
  December 8-14, 2019, Vancouver, BC, Canada}, pages 3261--3275.

\bibitem[{Wang et~al.(2019{\natexlab{b}})Wang, Singh, Michael, Hill, Levy, and
  Bowman}]{wang2018glue}
Alex Wang, Amanpreet Singh, Julian Michael, Felix Hill, Omer Levy, and
  Samuel~R. Bowman. 2019{\natexlab{b}}.
\newblock \href {https://openreview.net/forum?id=rJ4km2R5t7} {{GLUE:} {A}
  multi-task benchmark and analysis platform for natural language
  understanding}.
\newblock In \emph{7th International Conference on Learning Representations,
  {ICLR} 2019, New Orleans, LA, USA, May 6-9, 2019}. OpenReview.net.

\bibitem[{Wolf et~al.(2020)Wolf, Debut, Sanh, Chaumond, Delangue, Moi, Cistac,
  Rault, Louf, Funtowicz, Davison, Shleifer, von Platen, Ma, Jernite, Plu, Xu,
  Le~Scao, Gugger, Drame, Lhoest, and Rush}]{wolf-etal-2020-transformers}
Thomas Wolf, Lysandre Debut, Victor Sanh, Julien Chaumond, Clement Delangue,
  Anthony Moi, Pierric Cistac, Tim Rault, Remi Louf, Morgan Funtowicz, Joe
  Davison, Sam Shleifer, Patrick von Platen, Clara Ma, Yacine Jernite, Julien
  Plu, Canwen Xu, Teven Le~Scao, Sylvain Gugger, Mariama Drame, Quentin Lhoest,
  and Alexander Rush. 2020.
\newblock \href {https://doi.org/10.18653/v1/2020.emnlp-demos.6} {Transformers:
  State-of-the-art natural language processing}.
\newblock In \emph{Proceedings of the 2020 Conference on Empirical Methods in
  Natural Language Processing: System Demonstrations}, pages 38--45, Online.
  Association for Computational Linguistics.

\bibitem[{Zhang et~al.(2020)Zhang, Zhao, Duan, Chen, Zhang, and
  Tang}]{zhang2020multi}
Dong Zhang, Shu Zhao, Zhen Duan, Jie Chen, Yanping Zhang, and Jie Tang. 2020.
\newblock A multi-label classification method using a hierarchical and
  transparent representation for paper-reviewer recommendation.
\newblock \emph{ACM Transactions on Information Systems (TOIS)}, 38(1):1--20.

\end{thebibliography}
\bibliographystyle{acl_natbib}

\clearpage
\newpage
\appendix

\section{Ablation Studies for SPECTER Single Domain}

We conduct another ablation study on \dataname using single domain (CS) SPECTER training. We report the results in \Cref{tab:ablation_single_domain} and observe the similar trend as in \Cref{tab:ablation}, which further supports our conclusion in the main paper.

\begin{table}[h!]
    \begin{adjustbox}{width=\columnwidth}
    \begin{tabular}{p{0.8\linewidth}ccc}
                                                             & \textbf{MAG} & \textbf{cite}      & \textbf{co-cite} \\
                                                             & \msa{F1}     & \msa{MAP}          & \msa{MAP} \\
        \midrule
        {\name (3 CLS, \(\lambda=0.1\))}                       & 80.24        & 81.57              & 69.12                      \\
        \specialrule{.2em}{.3em}{.3em}
        \textbf{CLS Embedding Independence} & & \\
        \multirow{2}{*}{\(\rightarrow \lambda=0.5\)}         & 79.94         & 81.33              & 68.46 \\
                                                             & \mdo{-1.52\%} & \mdo{-1.33\%}      & \mdo{-2.15\%}\\
        \midrule
        \multirow{2}{*}{\(\rightarrow \lambda=0.0\)}         & 80.18         & 81.39              & 68.95 \\
                                                             & \mdo{-0.28\%} & \mdo{-0.98\%}      & \mdo{-0.55\%}\\
        \midrule
        \multirow{2}{*}{\(\rightarrow \lambda=1.0\)}         & 79.39         & 79.85              & 66.44 \\
                                                             & \mdo{-4.30\%} & \mdo{-9.35\%}      & \mdo{-8.68\%}\\
        \specialrule{.15em}{.3em}{.3em}
        \textbf{Architectural Changes} & & \\
        \multirow{2}{*}{\(\rightarrow\) 1 CLS token}        & 79.98 & 81.09 & 67.84 \\
                                                             & \mdo{-1.29\%} & \mdo{-2.62\%}      & \mdo{-4.16\%}\\
        \midrule
        \multirow{2}{*}{\(\rightarrow\) 5 CLS tokens}        & 80.14         & 80.78              & 68.03 \\
                                                             & \mdo{-0.48\%} & \mdo{-4.28\%}      & \mdo{-3.54\%}\\
        \midrule
        \multirow{2}{*}{\(\rightarrow\) No linear layer injection in BERT}   & 80.09         & 80.70            & 67.67 \\
                                                                             & \mdo{-0.73\%} & \mdo{-4.72\%}    & \mdo{-4.69\%}\\
        \midrule
        \multirow{2}{*}{\(\rightarrow\) No re-parameterization trick}        & 80.05         & 81.03            & 67.67 \\
                                                                             & \mdo{-0.92\%} & \mdo{-2.96\%}    & \mdo{-3.94\%}\\
        \bottomrule
    \end{tabular}
    \end{adjustbox}
    \caption{Results from ablation studies conducted on SPECTER and single domain (CS) training data. All scores are averaged over four random seeds. Percentages indicate relative error reduction over the baseline (3 CLS, \(\lambda=0.1\)).}
    \label{tab:ablation_single_domain}
\end{table}

\section{Comparison on SciDocs}

We report the SciDocs performance of our models and baselines in \Cref{tab:results_original_scidocs}. In all SPECTER training sets, \name consistently and significantly outperforms the SPECTER baseline.
For example, in \Cref{tab:results_original_scidocs} (b), we train the encoders using medicine papers and test them using mostly using CS papers. The large improvement of \name demonstrates its strong out-of-domain generalization ability.

In all SciNCL training sets, \name performs similarly compared to the single embedding baseline. One possible reason is that SciNCL fine-tunes their sampling hyperparameters to optimize the performance of SciDocs while we did not conduct such fine-tuning for \name. Another possible reason is that SciNCL tends to remove cross-domain citations in the training set as we mentioned in the main paper. How to better combine multiple CLS tokens and SciNCL would be an interesting future research direction.

Notice that in \Cref{tab:results_original_scidocs} (a), our baseline is slightly better than the reported SPECTER performance in \citet{cohan2020specter} mainly because we use the pyTorch version of SPECTER rather than the original AllenNLP version. In \Cref{tab:results_original_scidocs} (c), our baseline is slightly worse than the reported SciNCL performance in~\citet{ostendorff2022neighborhood}. We hypothesize that this is because we directly run the SPECTER's training code on the training data prepared by SciNCL rather than directly using SciNCL's training code. 

The ablation studies in \Cref{tab:results_original_scidocs} (d) suggest that the SciDocs score is not very sensitive to $\lambda$ values; 5 CLS performs slightly better than 3 CLS; Using 1 CLS, removing linear layers, or re-parameterization still significantly degrades the performance.

\begin{table*}[t]
    \centering
    \begin{subtable}[t]{\textwidth}
    \begin{adjustbox}{width=\textwidth}
    \begin{tabular}{c|c|c|cc|cc|cc|cc|cc|c}
                                             & \textbf{MAG} & \textbf{MeSH} & \multicolumn{2}{c|}{\textbf{co-view}} & \multicolumn{2}{c|}{\textbf{co-read}} & \multicolumn{2}{c|}{\textbf{cite}} & \multicolumn{2}{c|}{\textbf{co-cite}} & \multicolumn{2}{c|}{\textbf{recomm}} & \multirow{2}{*}{\textbf{Avg}}\\
                                             & \msa{F1} & \msa{F1}     & \msa{MAP}           & \msa{nDCG}         & \msa{MAP}          & \msa{nDCG} & \msa{MAP}          & \msa{nDCG}  & \msa{MAP}          & \msa{nDCG}   & \msa{nDCG}          & \msa{P@1}              \\
        \midrule
        {\centering SPECTER*}                 & 82 & 86.4 & 83.6 & 91.5 & 84.5 & 92.4 & 88.3 & 94.9 & 88.1 & 94.8 & 53.9 & 20.0 & 80.03 \\
        \midrule
        {\centering SPECTER}                 & 78.90        & 86.41          & 83.56 & 91.49      & 84.83 & 92.57    & 92.40 & 96.82 & 88.91 & 95.14 & 54.22 & 21.03 & 80.52 \\
        \hdashline
        {\centering \name}                   & 80.24        & 86.94          & 83.91        & 91.62        & 85.61        & 92.96    & 94.09 & 97.54 & 89.40 & 95.38 & 54.49 & 20.62 & 81.06 \\
        {(3 CLS, \(\lambda=0.1\))}           & \mup{6.35\%} & \mup{3.92\%}   & \mup{2.14\%} & \mup{1.59\%} & \mup{5.09\%} & \mup{5.15\%} & \mup{22.15\%} & \mup{22.72\%} & \mup{4.39\%} & \mup{4.94\%} & \mup{0.57\%} &\mdo{-0.52\%} & \mup{2.78\%} \\
        \specialrule{.2em}{.3em}{.3em}
        {\centering SciNCL*}    & 81.4 & 88.7 & 85.3 & 92.3 & 87.5 & 93.9 & 93.6 & 97.3 & 91.6 & 96.4 & 53.9 & 19.3 & 81.77 \\
        \midrule
        {\centering SciNCL}                 & 79.59 & 88.63 & 85.05 & 92.18 & 87.13 & 93.72 & 93.45 & 97.27 & 91.68 & 96.46 & 55.03 & 21.09 & 81.77 \\
        \hdashline
        {\centering \name}                   & 80.73 & 88.51 & 85.02 & 92.18 & 87.36 & 93.84 & 92.73 & 96.90 & 91.46 & 96.33 & 54.69 & 20.32 & 81.67 \\
        {(3 CLS, \(\lambda=0.1\))}           & \mup{5.58\%} & \mdo{-1.06\%} & \mdo{-0.18\%} & \mup{0.00\%} & \mup{1.77\%} & \mup{1.83\%} & \mdo{-11.03\%} & \mdo{-13.63\%} & \mdo{-2.61\%} & \mdo{-3.74\%} & \mdo{-0.75\%} & \mdo{-0.98\%} & \mdo{-0.56\%}\\
        \bottomrule
    \end{tabular}
    \end{adjustbox}
    \caption{Single domain (CS) training}
    \end{subtable}
    \hspace{\fill}
        \begin{subtable}[t]{\textwidth}
    \begin{adjustbox}{width=\textwidth}
    \begin{tabular}{c|c|c|cc|cc|cc|cc|cc|c}
                                             & \textbf{MAG} & \textbf{MeSH} & \multicolumn{2}{c|}{\textbf{co-view}} & \multicolumn{2}{c|}{\textbf{co-read}} & \multicolumn{2}{c|}{\textbf{cite}} & \multicolumn{2}{c|}{\textbf{co-cite}} & \multicolumn{2}{c|}{\textbf{recomm}} & \multirow{2}{*}{\textbf{Avg}}\\
                                             & \msa{F1} & \msa{F1}     & \msa{MAP}           & \msa{nDCG}         & \msa{MAP}          & \msa{nDCG} & \msa{MAP}          & \msa{nDCG}  & \msa{MAP}          & \msa{nDCG}   & \msa{nDCG}          & \msa{P@1}              \\
        \midrule
        {\centering SPECTER}                 & 79.70 & 87.38 & 81.23 & 90.34 & 81.60 & 90.95 & 81.59 & 91.54 & 85.78 & 93.72 & 54.55 & 20.49 & 78.24 \\
        \hdashline
        {\centering \name}                   & 81.14 & 88.15 & 82.65 & 90.93 & 83.40 & 91.88 & 86.52 & 94.08 & 87.66 & 94.60 & 54.74 & 20.78 & 79.71 \\
        {(3 CLS, \(\lambda=0.1\))}           & \mup{7.10\%} & \mup{6.12\%} & \mup{7.53\%} & \mup{6.16\%} & \mup{9.79\%} & \mup{10.28\%} & \mup{26.79\%} & \mup{30.01\%} & \mup{13.17\%} & \mup{14.04\%} & \mup{0.41\%} & \mup{0.37\%} & \mup{6.76\%} \\
        \bottomrule
    \end{tabular}
    \end{adjustbox}
    \caption{Single domain (medicine) training}
    \end{subtable}
    \hspace{\fill}
    \begin{subtable}[t]{\textwidth}
    \begin{adjustbox}{width=\textwidth}
    \begin{tabular}{c|c|c|cc|cc|cc|cc|cc|c}
                                             & \textbf{MAG} & \textbf{MeSH} & \multicolumn{2}{c|}{\textbf{co-view}} & \multicolumn{2}{c|}{\textbf{co-read}} & \multicolumn{2}{c|}{\textbf{cite}} & \multicolumn{2}{c|}{\textbf{co-cite}} & \multicolumn{2}{c|}{\textbf{recomm}} & \multirow{2}{*}{\textbf{Avg}}\\
                                             & \msa{F1} & \msa{F1}     & \msa{MAP}           & \msa{nDCG}         & \msa{MAP}          & \msa{nDCG} & \msa{MAP}          & \msa{nDCG}  & \msa{MAP}          & \msa{nDCG}   & \msa{nDCG}          & \msa{P@1}              \\
        \midrule
        {\centering SPECTER}                 & 79.99 & 87.79 & 81.56 & 90.45 & 81.90 & 91.15 & 82.29 & 92.00 & 85.90 & 93.73 & 54.06 & 19.63 & 78.37 \\
        \hdashline
        {\centering \name}                   & 81.36 & 88.35 & 82.85 & 91.08 & 83.87 & 92.16 & 88.16 & 94.86 & 88.14 & 94.86 & 54.13 & 19.84 & 79.97 \\
        {(3 CLS, \(\lambda=0.1\))}           & \mup{6.85\%} & \mup{4.53\%} & \mup{7.03\%} & \mup{6.60\%} & \mup{10.89\%} & \mup{11.39\%} & \mup{33.12\%} & \mup{35.76\%} & \mup{15.85\%} & \mup{18.01\%} & \mup{0.16\%} & \mup{0.26\%}  & \mup{7.40\%} \\
        \specialrule{.2em}{.3em}{.3em}
        {\centering SciNCL*}                 & 81.3 & 89.4 & 84.3 & 91.8 & 85.6 & 92.8 & 91.4 & 96.3 & 90.1 & 95.7 & 54.3 & 19.9 & 81.08 \\
        \midrule
        {\centering SciNCL}                 & 80.27 & 88.78 & 83.79 & 91.53 & 84.94 & 92.52 & 90.78 & 96.00 & 89.61 & 95.48 & 54.68 & 20.49 & 80.74 \\
        \hdashline
        {\centering \name}                   & 81.04 & 89.34 & 83.73 & 91.47 & 85.27 & 92.73 & 90.16 & 95.74 & 89.82 & 95.56 & 55.00 & 20.97 & 80.9 \\
        {(3 CLS, \(\lambda=0.1\))}           & \mup{3.90\%} & \mup{4.97\%} & \mdo{-0.40\%} & \mdo{-0.77\%} & \mup{2.21\%} & \mup{2.84\%} & \mdo{-6.75\%} & \mdo{-6.69\%} & \mup{2.00\%} & \mup{1.77\%} & \mup{0.71\%} & \mup{0.61\%} &  \mup{0.84\%} \\
        \bottomrule
    \end{tabular}
    \end{adjustbox}
    \caption{Multiple domain training}
    \end{subtable}
    \hspace{\fill}
    \begin{subtable}[t]{\textwidth}
    \begin{adjustbox}{width=\textwidth}
    \begin{tabular}{c|c|c|cc|cc|cc|cc|cc|c}
                                             & \textbf{MAG} & \textbf{MeSH} & \multicolumn{2}{c|}{\textbf{co-view}} & \multicolumn{2}{c|}{\textbf{co-read}} & \multicolumn{2}{c|}{\textbf{cite}} & \multicolumn{2}{c|}{\textbf{co-cite}} & \multicolumn{2}{c|}{\textbf{recomm}} & \multirow{2}{*}{\textbf{Avg}}\\
                                             & \msa{F1} & \msa{F1}     & \msa{MAP}           & \msa{nDCG}         & \msa{MAP}          & \msa{nDCG} & \msa{MAP}          & \msa{nDCG}  & \msa{MAP}          & \msa{nDCG}   & \msa{nDCG}          & \msa{P@1}              \\
        \midrule
        {\centering 3 CLS, \(\lambda=0.1\)}                   & 81.36 & 88.35 & 82.85 & 91.08 & 83.87 & 92.16 & 88.16 & 94.86 & 88.14 & 94.86 & 54.13 & 19.84 & 79.97 \\
        \hdashline
        {\centering 3 CLS, \(\lambda=0\)}                   & 81.06 & 88.22 & 82.78 & 91.04 & 83.94 & 92.18 & 88.13 & 94.83 & 87.96 & 94.80 & 54.19 & 20.16 & 79.94 \\
        & \mdo{-1.64\%} & \mdo{-1.12\%} & \mdo{-0.41\%} & \mdo{-0.48\%} & \mup{0.40\%} & \mup{0.19\%} & \mdo{-0.23\%} & \mdo{-0.53\%} & \mdo{-1.50\%} & \mdo{-1.17\%} & \mup{0.13\%} & \mup{0.40\%} & \mdo{-0.16\%} \\
        \hdashline
        {\centering 3 CLS, \(\lambda=0.5\)}                   & 80.70 & 87.85 & 82.96 & 91.16 & 83.91 & 92.12 & 87.55 & 94.55 & 88.22 & 94.87 & 54.37 & 20.48 & 79.90 \\
        & \mdo{-3.53\%} & \mdo{-4.29\%} & \mup{0.64\%} & \mup{0.81\%} & \mup{0.25\%} & \mdo{-0.48\%} & \mdo{-5.13\%} & \mdo{-5.98\%} & \mup{0.67\%} & \mup{0.29\%} & \mup{0.52\%} & \mup{0.80\%} & \mdo{-0.38\%}  \\
        \hdashline
        {\centering 3 CLS, \(\lambda=1\)}                   & 79.78 & 86.94 & 82.87 & 91.10 & 84.08 & 92.21 & 90.39 & 95.87 & 87.96 & 94.76 & 54.00 & 20.32 & 80.02 \\
        & \mdo{-8.49\%} & \mdo{-12.05\%} & \mup{0.09\%} & \mup{0.17\%} & \mup{1.27\%} & \mup{0.67\%} & \mup{18.83\%} & \mup{19.64\%} & \mdo{-1.54\%} & \mdo{-1.90\%} & \mdo{-0.28\%} & \mup{0.60\%} & \mup{0.25\%} \\
        \hdashline
        {\centering 1 CLS}                   & 80.32 & 87.99 & 82.24 & 90.74 & 83.30 & 91.93 & 88.23 & 94.88 & 87.04 & 94.32 & 53.90 & 19.77 & 79.55 \\
        & \mdo{-5.57\%} & \mdo{-3.02\%} & \mdo{-3.57\%} & \mdo{-3.84\%} & \mdo{-3.55\%} & \mdo{-3.00\%} & \mup{0.63\%} & \mup{0.44\%} & \mdo{-9.29\%} & \mdo{-10.40\%} & \mdo{-0.52\%} & \mdo{-0.08\%} & \mdo{-2.08\%}  \\
        \hdashline
        {\centering 5 CLS, \(\lambda=0.1\)}                   & 80.83 & 88.79 & 83.48 & 91.41 & 84.46 & 92.41 & 88.15 & 94.78 & 88.72 & 95.09 & 54.29 & 20.36 & 80.23 \\
        & \mdo{-2.84\%} & \mup{3.78\%} & \mup{3.64\%} & \mup{3.67\%} & \mup{3.66\%} & \mup{3.16\%} & \mdo{-0.06\%} & \mdo{-1.51\%} & \mup{4.91\%} & \mup{4.62\%} & \mup{0.35\%} & \mup{0.65\%} & \mup{1.29\%} \\
        \hdashline
        {\centering 3 CLS, \(\lambda=0.1\)} & 80.82 & 87.79 & 82.50 & 90.82 & 83.31 & 91.90 & 87.37 & 94.51 & 87.30 & 94.40 & 53.79 & 19.11 & 79.47 \\
        {\centering w/o linear layers} & \mdo{-2.88\%} & \mdo{-4.80\%} & \mdo{-2.08\%} & \mdo{-2.92\%} & \mdo{-3.49\%} & \mdo{-3.35\%} & \mdo{-6.69\%} & \mdo{-6.85\%} & \mdo{-7.10\%} & \mdo{-8.84\%} & \mdo{-0.75\%} & \mdo{-0.92\%} & \mdo{-2.52\%} \\
        \hdashline
        {\centering 3 CLS, \(\lambda=0.1\)} & 80.89 & 87.82 & 82.09 & 90.67 & 82.92 & 91.67 & 88.79 & 95.18 & 86.86 & 94.27 & 54.44 & 20.89 & 79.71 \\
        {\centering w/o reparameterization} & \mdo{-2.55\%} & \mdo{-4.50\%} & \mdo{-4.43\%} & \mdo{-4.65\%} & \mdo{-5.91\%} & \mdo{-6.28\%} & \mup{5.34\%} & \mup{6.32\%} & \mdo{-10.81\%} & \mdo{-11.32\%} & \mup{0.67\%} & \mup{1.31\%} & \mdo{-1.32\%} \\
        \bottomrule
    \end{tabular}
    \end{adjustbox}
    \caption{Ablation study (multiple domain SPECTER training)}
    \end{subtable}
    \caption{Results of our methods and baselines. All scores are averaged over four random seeds. Percentages indicate relative error reduction over the baseline. In the (d) ablation study section, the more negative error reduction {\color{redbw} (red)} is better, while more positive error reduction {\color{greenbw} (green)} is better in other sections. *The scores come from the original papers.}
    \label{tab:results_original_scidocs}
\end{table*}

\section{Training and Testing Dataset Information}
\label{sec:multidomain_pretraining}

\subsection{Creating training data}

\paragraph{SPECTER, single domain (CS)} We use the original dataset files provided by the SPECTER authors.\footnote{\url{https://github.com/allenai/specter/issues/2\#issuecomment-625428992}}

\paragraph{SPECTER, multi domain} We use the papers in shard 11 of the 20200705 version of S2ORC \citep{lo2020s2orc} as query papers. Since we tried to be unbiased as possible towards different subject areas during the paper selection, we filter out any papers without MAG information in S2ORC. Then we scan the entirety of S2ORC to fetch all direct and indirect citations, as defined in \citet{cohan2020specter}, for the chosen query papers. We feed the acquired citation information into the sampling code provided by the SPECTER authors.

\paragraph{SPECTER, single domain (Medicine)} We follow the same steps as SPECTER multi-domain, except for the use of shard 22, 33, and 44 and limiting all query papers to belong to the `Medicine' MAG field.

\paragraph{SciNCL} For SciNCL single domain, we use the `SciNCL w/ leakage' dataset.\footnote{\url{https://github.com/malteos/scincl/releases/tag/0.1}} For SciNCL multi-domain, we use the `SciNCL w/o leakage' dataset.\footnote{\url{https://github.com/malteos/scincl/releases/tag/0.1-wol}} We examined the `SciNCL w/o leakage' dataset and found it to be fairly balanced in terms of subject field distribution of query papers, as they also randomly choose query papers from S2ORC.

\subsection{Creating \dataname}

For a better measurement of multi-domain performance, we have created the multi-domain (co-)citation prediction tasks. We refer to the collection of 3 multi-domain tasks, \textsf{multi. cite}, \textsf{multi. co-cite}, and \textsf{MAG} as \dataname.

For \textsf{multi. cite} and \textsf{multi. co-cite} datasets, we use shard 7 of the 20200705 version of S2ORC~\citep{lo2020s2orc} as query papers, avoiding a certain domain from being the majority of query papers. For each query, we collect 500 negative papers and up to 5 positive papers. This is the same setting used in the original SciDocs \citep{cohan2020specter}. The task is to assign higher similarity scores to the positive papers and lower scores to the negative papers. In both datasets, the negative samples come from randomly sampled papers. In the \textsf{multi. cite} dataset, the positive samples are the papers cited by the query paper. In the \textsf{multi. co-cite} dataset, the positive samples and the query paper are both cited by another paper. We made minimal modification to the SciDocs execution code\footnote{\url{https://github.com/allenai/scidocs}} to allow testing with our custom dataset files.

For \textsf{MAG}, we use the same test set included in the original SciDocs.

\subsection{MAG subject field distribution of training and testing datasets}
\label{sec:dataset_stat}

Please refer to \Cref{tab:dataset_stat}.

\begin{table*}[ht]
    \centering
    \begin{adjustbox}{width=\textwidth}
    \begin{tabular}{cccccccccc}
        \toprule
        & \multicolumn{4}{c}{\textit{Training}} & \multicolumn{2}{c}{\textit{Testing - SciDocs}} & \multicolumn{3}{c}{\textit{Testing - }\dataname} \\
        \hdashline
        \multirow{3}{*}{\textbf{MAG field}} & \textbf{SPECTER}        & \textbf{SciNCL}  & \textbf{SPECTER$_{\text{multi}}$} & \textbf{SciNCL$_{\text{multi}}$} & \textbf{cite} & \textbf{cocite} & \textbf{MAG} & \textbf{multi. cite} & \textbf{multi. cocite}   \\

        & \msa{Percentage} & \msa{Percentage}   & \msa{Percentage} & \msa{Percentage} & \msa{Percentage} & \msa{Percentage} & \msa{Percentage} & \msa{Percentage} & \msa{Percentage} \\
        & \msa{(Count)}  & \msa{(Count)} & \msa{(Count)} & \msa{(Count)} & \msa{(Count)} & \msa{(Count)} & \msa{(Count)} & \msa{(Count)} & \msa{(Count)} \\
        \midrule
        \multirow{2}{*}{Art}               & 0.03\% & 0.04\% & 0.02\% & 0.06\% & 0\% & 0\% & 4.67\% & 0.03\% & 0\% \\
                                           & \msa{(53)} & \msa{(62)} & \msa{(12)} & \msa{(108)} & \msa{(0)} & \msa{(0)} & \msa{(175)} & \msa{(13)} & \msa{(0)}  \\
        \midrule
        \multirow{2}{*}{Biology}          & 1.03\% & 1.31\% & 19.55\% & 13.77\% & 0.49\% & 0.33\% & 6.05\% & 19.66\% & 25.14\% \\
                                        & \msa{(1748)} & \msa{(2257)} & \msa{(14561)} & \msa{(23904)} & \msa{(6)} & \msa{(4)} & \msa{(227)} & \msa{(9813)} & \msa{(785)} \\
        \midrule
        \multirow{2}{*}{Business}          & 0.68\% & 0.67\% & 0.58\% & 0.92\% & 0.58\% & 0.41\% & 5.23\% & 0.58\% & 0.29\% \\
                                           & \msa{(1156)} & \msa{(1148)} & \msa{(433)} & \msa{(1591)} & \msa{(7)} & \msa{(5)} & \msa{(196)} & \msa{(289)} & \msa{(9)}  \\
        \midrule
        \multirow{2}{*}{Chemistry}          & 0.17\% & 0.29\% & 3.27\% & 3.65\% & 0.16\% & 0.08\% & 5.25\% & 3.54\% & 2.59\% \\
                                           & \msa{(282)} & \msa{(495)} & \msa{(2433)} & \msa{(6336)} & \msa{(2)} & \msa{(1)} & \msa{(197)} & \msa{(1767)} & \msa{(81)}  \\
        \midrule
        \multirow{2}{*}{Computer Science}  & \textbf{61.00\%} & \textbf{61.78\%} & 12.03\% & 15.20\% & \textbf{65.16\%} & \textbf{62.78\%} & 5.07\% & 11.61\% & 10.63\% \\
                                           & \msa{(103869)} & \msa{(106268)} & \msa{(8962)} & \msa{(26391)} & \msa{(791)} & \msa{(769)} & \msa{(190)} & \msa{(5795)} & \msa{(332)}  \\
        \midrule
        \multirow{2}{*}{Economics}          & 0.63\% & 0.67\% & 1.82\% & 2.06\% & 0.66\% & 0.49\% & 5.76\% & 1.82\% & 1.19\% \\
                                           & \msa{(1078)} & \msa{(1159)} & \msa{(1355)} & \msa{(3578)} & \msa{(8)} & \msa{(6)} & \msa{(216)} & \msa{(907)} & \msa{(37)}  \\
        \midrule
        \multirow{2}{*}{Engineering}       & 7.17\% & 7.34\% & 2.13\% & 4.79\% & 6.59\% & 6.29\% & 5.55\% & 2.56\% & 1.31\% \\
                                           & \msa{(12212)} & \msa{(12623)} & \msa{(1586)} & \msa{(8315)} & \msa{(80)} & \msa{(77)} & \msa{(208)} & \msa{(1277)} & \msa{(41)}  \\
        \midrule
        \multirow{2}{*}{Environmental Science}       & 0.05\% & 0.06\% & 0.33\% & 0.47\% & 0\% & 0\% & 5.76\% & 0.37\% & 0.10\% \\
                                           & \msa{(85)} & \msa{(102)} & \msa{(243)} & \msa{(821)} & \msa{(0)} & \msa{(0)} & \msa{(216)} & \msa{(183)} & \msa{(3)}  \\
        \midrule
        \multirow{2}{*}{Geography}       & 0.24\% & 0.25\% & 0.47\% & 0.76\% & 0.41\% & 0.24\% & 4.59\% & 0.49\% & 0.22\% \\
                                           & \msa{(405)} & \msa{(430)} & \msa{(350)} & \msa{(1322)} & \msa{(5)} & \msa{(3)} & \msa{(172)} & \msa{(247)} & \msa{(7)}  \\
        \midrule
        \multirow{2}{*}{Geology}       & 0.11\% & 0.13\% & 1.32\% & 1.38\% & 0.08\% & 0.16\% & 5.65\% & 1.16\% & 0.64\% \\
                                           & \msa{(183)} & \msa{(216)} & \msa{(982)} & \msa{(2395)} & \msa{(1)} & \msa{(2)} & \msa{(212)} & \msa{(577)} & \msa{(20)}  \\
        \midrule
        \multirow{2}{*}{History}       & 0.01\% & 0.02\% & 0.02\% & 0.13\% & 0\% & 0\% & 4.99\% & 0.05\% & 0.10\% \\
                                           & \msa{(20)} & \msa{(26)} & \msa{(15)} & \msa{(223)} & \msa{(0)} & \msa{(0)} & \msa{(187)} & \msa{(24)} & \msa{(3)}  \\
        \midrule
        \multirow{2}{*}{Materials Science}       &0.44\% & 0.50\% & 1.31\% & 2.21\% & 0.33\% & 0.82\% & 5.76\% & 1.65\% & 0.70\% \\
                                           & \msa{(749)} & \msa{(867)} & \msa{(976)} & \msa{(3844)} & \msa{(4)} & \msa{(10)} & \msa{(216)} & \msa{(825)} & \msa{(22)} \\
        \midrule
        \multirow{2}{*}{Mathematics}       & 7.53\% & 7.84\% & 7.95\% & 5.75\% & 7.50\% & 9.06\% & 5.20\% & 8.01\% & 6.47\% \\
                                           & \msa{(12828)} & \msa{(13490)} & \msa{(5923)} & \msa{(9988)} & \msa{(91)} & \msa{(111)} & \msa{(195)} & \msa{(3997)} & \msa{(202)}  \\
        \midrule
        \multirow{2}{*}{Medicine}       & 8.59\% & 9.41\% & 40.72\% & 31.79\% & 6.10\% & 7.67\% & 4.88\% & 39.37\% & 43.27\% \\
                                           & \msa{(14629)} & \msa{(16187)} & \msa{(30325)} & \msa{(55178)} & \msa{(74)} & \msa{(94)} & \msa{(183)} & \msa{(19649)} & \msa{(1351)}  \\
        \midrule
        \multirow{2}{*}{Philosophy}       & 0.02\% & 0.03\% & 0.03\% & 0.10\% & 0\% & 0\% & 4.67\% & 0.05\% & 0\% \\
                                           & \msa{(41)} & \msa{(52)} & \msa{(19)} & \msa{(168)} & \msa{(0)} & \msa{(0)} & \msa{(175)} & \msa{(26)} & \msa{(0)}  \\
        \midrule
        \multirow{2}{*}{Physics}       & 1.41\% & 1.51\% & 3.93\% & 3.08\% & 1.32\% & 2.69\% & 5.25\% & 4.49\% & 2.75\% \\
                                           & \msa{(2399)} & \msa{(2605)} & \msa{(2930)} & \msa{(5353)} & \msa{(16)} & \msa{(33)} & \msa{(197)} & \msa{(2242)} & \msa{(86)}  \\
        \midrule
        \multirow{2}{*}{Political Science}       & 0.13\% & 0.14\% & 0.22\% & 0.61\% & 0.16\% & 0.08\% & 5.36\% & 0.25\% & 0.03\% \\
                                           & \msa{(219)} & \msa{(240)} & \msa{(165)} & \msa{(1054)} & \msa{(2)} & \msa{(1)} & \msa{(201)} & \msa{(123)} & \msa{(1)}  \\
        \midrule
        \multirow{2}{*}{Psychology}       & 4.18\% & 4.27\% & 3.79\% & 3.57\% & 3.13\% & 3.27\% & 5.41\% & 3.77\% & 4.42\% \\
                                           & \msa{(7125)} & \msa{(7340)} & \msa{(2819)} & \msa{(6197)} & \msa{(38)} & \msa{(40)} & \msa{(203)} & \msa{(1884)} & \msa{(138)}  \\
        \midrule
        \multirow{2}{*}{Sociology}       & 0.34\% & 0.37\% & 0.52\% & 0.94\% & 0.25\% & 0.16\% & 4.93\% & 0.55\% & 0.13\% \\
                                           & \msa{(576)} & \msa{(641)} & \msa{(389)} & \msa{(1638)} & \msa{(3)} & \msa{(2)} & \msa{(185)} & \msa{(276)} & \msa{(4)}  \\
        \midrule
        \multirow{2}{*}{Unknown}               & 6.23\% & 3.38\% & 0.00\% & 8.74\% & 7.08\% & 5.47\% & 0\% & 0\% & 0\% \\
                                           & \msa{(10611)} & \msa{(5811)} & \msa{(0)} & \msa{(15169)} & \msa{(86)} & \msa{(67)} & \msa{(0)} & \msa{(0)} & \msa{(0)}  \\
        \specialrule{.2em}{.3em}{.3em}
        \multirow{2}{*}{Total}       & 100.00\% & 100.00\% & 100.00\% & 100.00\% & 100.00\% & 100.00\% & 100.00\% & 100.00\% & 100.00\% \\
                                           & \msa{(170268)} & \msa{(172019)} & \msa{(74478)} & \msa{(173573)} & \msa{(1214)} & \msa{(1225)} & \msa{(3751)} & \msa{(49914)} & \msa{(3122)}  \\
        \bottomrule
    \end{tabular}
    \end{adjustbox}
    \caption{Training and testing dataset statistics. Note: Since some papers are categorized under multiple MAG fields in S2ORC, they are counted more than once in this table. Unknown refers to the papers without MAG information in S2ORC.}
    \label{tab:dataset_stat}
\end{table*}
\section{Reproducibility Information}
\label{appendix:other_training}

All the program codes used to produce results presented in this paper are available at \url{https://link.iamblogger.net/multi2spe}.

\paragraph{Implementation of architectural changes} For multiple CLS tokens, we use \texttt{[unused]} tokens available in SciBERT vocabulary except for the first CLS token, where we make use of the existing \texttt{[CLS]} token. For the linear layers we add to the BERT architecture, the ones after 4th and 8th are initialized with identity matrix to allow pretrained weights to output correctly in the beginning. The ones after 12th are randomly initialized. We note that our current best model, \name (3 CLS, \(\lambda=0.1\)) have 114.8M parameters. which is relatively a small increase from the 109M parameters of BERT$_{\text{Base}}$ with sequence classification head. Similarly to Multi-CLS BERT~\citep{MultiCLSBert}, 
We insert additional linear transformations after the 4th, 8th, and 12th BERT layers.

\paragraph{Training hyperparameters} Since we compare the performance of our models with the SPECTER/SciNCL baselines, we replicate the training setup originally used in SPECTER as much as possible. We truncate the abstract if the length of title and abstract is larger than 512. We initialize the BERT layer with the pretrained SciBERT (\texttt{scibert\_scivocab\_uncased}.)\footnote{\url{https://huggingface.co/allenai/scibert\_scivocab\_uncased}} We use the Adam optimizer with a learning rate of 2e-5. A linear learning rate scheduler with warm up fraction of 0.1 is used. Since SPECTER uses an effective batch size of 32 with gradient accumulation, we achieve the same effective batch size by doing gradient accumulation at every 16 steps with the real batch size of 2. We train all the models for 2 epochs.

\paragraph{Random seeds} All experiments are ran with 4 different seeds, \texttt{1783, 1918, 1945, 1991.} All the reported metrics in this paper are the average scores from the 4 seeds, unless otherwise noted. 

\paragraph{Hardwares and softwares used} To fully utilize all the GPU resources available to us, We trained all our models using multiple NVIDIA RTX 2080 Ti, GTX 1080 Ti and GTX TITAN X GPUs. To achieve better reproducibility, each random seed was always ran with the same GPU model (\texttt{1783} \(\rightarrow\) TITAN X, \texttt{1918} \(\rightarrow\) 2080 Ti, \texttt{1945} \(\rightarrow\) 1080 Ti, \texttt{1991} \(\rightarrow\) 2080 Ti.)

We use the version 4.9.2 of HuggingFace Transformers library \citep{wolf-etal-2020-transformers} for BERT implementation, and PyTorch Lightning version 1.4.2 \citep{william_falcon_2020_3828935} alongside PyTorch 1.8.2 \citep{NEURIPS2019_9015} for training logic organization.

We also make use of the source codes released by the authors of SPECTER\footnote{\url{https://github.com/allenai/specter}} and SciNCL\footnote{\url{https://github.com/malteos/scincl}} to perform dataset creation and loading.

\end{document}